\documentclass[letterpaper]{article} 
\usepackage{aaai22}  
\usepackage{times}  
\usepackage{helvet}  
\usepackage{courier}  
\usepackage[hyphens]{url}  
\usepackage{graphicx} 
\usepackage{natbib}  
\usepackage{caption} 
\DeclareCaptionStyle{ruled}{labelfont=normalfont,labelsep=colon,strut=off} 
\usepackage{tabularx}
\usepackage{xparse}
\usepackage{amsmath}
\usepackage{amsthm}
\usepackage{amssymb}
\usepackage{xspace}
\usepackage{relsize}
\usepackage{multirow}
\usepackage{comment}
\usepackage{xcolor}
\usepackage{diagbox}
\usepackage{adjustbox}
\usepackage{pdflscape}

\usepackage{amsmath,amsfonts,bm}

\def\eqref#1{equation~\ref{#1}}

\def\1{\bm{1}}
\def\0{\bm{0}}

\usepackage{alphabeta}

\def\vmu{{\bm{\mu}}}

\def\vc{{\bm{c}}}

\def\vl{{\bm{l}}}

\def\vx{{\bm{x}}}

\def\vz{{\bm{z}}}

\def\evmu{{{\mu}}}

\def\evz{{z}}

\DeclareMathAlphabet{\mathsfit}{\encodingdefault}{\sfdefault}{m}{sl}
\SetMathAlphabet{\mathsfit}{bold}{\encodingdefault}{\sfdefault}{bx}{n}

\newcommand{\E}{\mathbb{E}}

\newcommand{\R}{\mathbb{R}}
\newcommand{\B}{\braces{0,1}}

\newcommand{\sigmoid}{\textsc{sigmoid}}

\newcommand{\KL}{D_{\mathrm{KL}}}

\DeclareMathOperator*{\argmax}{arg\,max}
\DeclareMathOperator*{\argmin}{arg\,min}

\def\Mid{\mathrel{\|}}

\newcommand{\brackets}[1]{{\left<#1\right>}}
\newcommand{\braces}[1]{{\left\{#1\right\}}}
\newcommand{\parens}[1]{{\left(#1\right)}}

\newcommand{\satisfies}{\models}

\renewcommand{\to}{\rightarrow}
\newcommand{\from}{\leftarrow}

\newcommand{\bern}{\text{\small Bern}}

\newcommand{\N}{\mathcal{N}}

\usepackage{algorithm}
\usepackage[noend]{algpseudocode} 

\usepackage{booktabs}   

\usepackage{stackengine}
\stackMath
\newcommand\tsup[2][2]{%
 \def\useanchorwidth{T}%
  \ifnum#1>1%
    \stackon[-.5pt]{\tsup[\numexpr#1-1\relax]{#2}}{\scriptscriptstyle\sim}%
  \else%
    \stackon[.5pt]{#2}{\scriptscriptstyle\sim}%
  \fi%
}

\newcommand{\function}[1]{\textsc{#1}}

\def\_{\\[-0.3em]}

\makeatletter
\let\@myref\ref

\newcommand{\refsec}[1]{Sec.\,\@myref{#1}}
\newcommand{\refseq}[1]{Sec.\,\@myref{#1}}
\newcommand{\refig}[1]{Fig.\,\@myref{#1}}
\newcommand{\reftbl}[1]{Table \@myref{#1}}
\newcommand{\refstep}[1]{Step \@myref{#1}}
\newcommand{\refalgo}[1]{Alg.\,\@myref{#1}}
\newcommand{\refchap}[1]{Chap.\,\@myref{#1}}
\newcommand{\reflst}[1]{List \@myref{#1}}
\newcommand{\refeq}[1]{Eq.\,\@myref{#1}}

\newcommand{\refthm}[1]{Thm.\,\@myref{#1}}
\newcommand{\refline}[1]{line\,\@myref{#1}}

\newcommand{\refsecs}[2]{Sec.\,\@myref{#1}-\@myref{#2}}
\newcommand{\refseqs}[2]{Sec.\,\@myref{#1}-\@myref{#2}}
\newcommand{\refigs}[2]{Fig.\,\@myref{#1}-\@myref{#2}}
\newcommand{\reftbls}[2]{Tables \@myref{#1}-\@myref{#2}}
\newcommand{\refsteps}[2]{Steps \@myref{#1}-\@myref{#2}}
\newcommand{\refalgos}[2]{Alg.\,\@myref{#1}-\@myref{#2}}
\newcommand{\refchaps}[2]{Chap.\,\@myref{#1}-\@myref{#2}}
\newcommand{\reflsts}[2]{Lists \@myref{#1}-\@myref{#2}}
\newcommand{\refeqs}[2]{Eq.\,\@myref{#1}-\@myref{#2}}
\newcommand{\refpages}[2]{p.\pageref{#1}-\@myref{#2}}
\newcommand{\refthms}[2]{Thm.\,\@myref{#1}-\@myref{#2}}
\newcommand{\reflines}[2]{line\,\@myref{#1}-\@myref{#2}}

\makeatother

\newcounter{list}[section]

\makeatletter
\@ifundefined{citet}{
  \NewDocumentCommand{\citet}{o m}{%
    \IfNoValueTF{#1}%
      {\citeauthor{#2} (\citeyear{#2})}
      {\citeauthor{#2} (\citeyear[#1]{#2})}%
  }
}{}
\@ifundefined{citep}{
  \NewDocumentCommand{\citep}{o m}{%
    \IfNoValueTF{#1}%
      {\cite{#2}}
      {\cite[#1]{#2}}%
  }
}{}
\makeatother

\newlength{\maxwidth}
\newcommand{\algalign}[2]
{\makebox[\maxwidth][r]{$#1{}$}${}#2$}

\newcommand{\encode}{\function{encode}}

\newcommand{\X}{\mathcal{X}}

\newcommand{\iw}{\function{IW}}
\newcommand{\db}{C}

\newcommand{\BC}{\function{bc}}

\makeatletter

\newcommand{\newheuristic}[2]{%
 \def#1{%
  \ifmmode%
  h^\text{#2}\xspace%
  \else%
  \text{#2}\xspace%
  \fi%
 }%
}

\newheuristic{\lmcut}{LMcut}
\newheuristic{\mands}{M\&S}
\newheuristic{\pdb}{PDB}
\newheuristic{\ff}{FF}
\newheuristic{\ce}{CEA}
\newheuristic{\cg}{CG}
\newheuristic{\ad}{add}
\newheuristic{\lc}{LC}

\newcommand{\newUnitCostHeuristic}[2]{%
 \def#1{%
  \ifmmode%
  \hat{h}^\text{#2}\xspace%
  \else%
  \text{#2}\xspace%
  \fi%
 }%
}

\newUnitCostHeuristic{\lmcuto}{LMcut}
\newUnitCostHeuristic{\mandso}{M\&S}
\newUnitCostHeuristic{\ffo}{FF}
\newUnitCostHeuristic{\ceo}{CEA}
\newUnitCostHeuristic{\cgo}{CG}
\newUnitCostHeuristic{\ado}{add}
\newUnitCostHeuristic{\gco}{GoalCount}
\newUnitCostHeuristic{\lco}{LC}

\makeatother

\author{Benjamin Ayton\textsuperscript{\rm 1}, Masataro Asai\textsuperscript{\rm 2}}
\title{Is Policy Learning Overrated?:\\ Width-Based Planning and Active Learning for Atari}
\affiliations {
Equal contributions.
\textsuperscript{\rm 1}MIT,
\textsuperscript{\rm 2}MIT-IBM Watson AI Lab\\
aytonb@mit.edu,masataro.asai@ibm.com
}

\begin{document}
\maketitle

\begin{abstract}
Width-based planning has shown promising results on Atari 2600 games using pixel input, while using
substantially fewer environment interactions than reinforcement learning. Recent width-based
approaches have computed feature vectors for each screen using a hand designed feature set or a
variational autoencoder trained on game screens (VAE-IW), and prune screens that do not have novel
features during the search.
We propose Olive (Online-VAE-IW), which updates the VAE features online
using active learning to maximize the utility of screens observed during planning.
Experimental results in 55 Atari games demonstrate that it
outperforms Rollout-IW by 42-to-11 and VAE-IW by 32-to-20.
Moreover, Olive outperforms existing work based on policy-learning
($\pi$-IW, DQN) trained with 100x training budget by 30-to-22 and 31-to-17,
and a state of the art data-efficient reinforcement learning (EfficientZero)
trained with the same training budget and ran with 1.8x planning budget by 18-to-7 in Atari 100k benchmark,
with no policy learning at all.
The source code is available at \url{github.com/ibm/atari-active-learning}.
\end{abstract}

\section{Introduction}
\label{sec:introduction}

Recent advancements in policy learning based on \emph{Reinforcement Learning} (RL)
\cite{sutton2018reinforcement} have made it possible to build an intelligent agent that operates
within a stochastic and noisy image-based interactive environment, which has been one of the major
goals of Artificial Intelligence.  The Arcade Learning Environment (ALE) \cite{bellemare2013arcade},
which allows access to pixel and memory features of Atari games, is a popular testbed for
testing modern RL approaches \cite{dqn}.  However, RL approaches are
notorious for their poor sample efficiency; they require a huge number of interactions with
the environment to learn the policy. This is especially true for sparse reward problems, such as
Montezuma's Revenge, where successful cases could require $10^{10}$ interactions with the
environment \cite{badia2020agent57}.  Poor sample efficiency hinders real-world applicability
when the number of interactions is restricted by practical constraints of non-simulated
environments, such as the runtime for collecting data and the safety of the environment.

Meanwhile, the \emph{planning setting} of the ALE provides an alternative benchmark where the focus
is placed on efficient search. Significant progress in this setting has been made on Iterative-Width
(IW) \cite{lipovetzky2012width,lipovetzky2015classical} recently, which prunes states if they do not
possess features that have not been seen during search, and has been demonstrated to outperform
Monte Carlo tree search based on UCT \cite{kocsis2006bandit}.  For example, p-IW
\cite{shleyfman2016blind} modifies IW to maximize the reward per feature, DASA
\cite{jinnai2017learning} learns and prunes actions that lead to identical states, Rollout-IW (RIW)
\cite{bandres2018planning} improves anytime characteristics of IW, which revolutionalized the field
with almost-real-time planning on pixel inputs, $\pi$-IW \cite{junyent2019deep} performs informed
search with a trainable policy represented by a neural network, and $\pi$-IW+
\cite{junyent2021hierarchical} also learns a value function.

Width-based approaches assume discrete inputs, such as the RAM state of Atari games,
but can work on continuous variables using discretization \cite{frances2017purely,teichteil2020boundary}.
$\pi$-IW discretizes the last hidden layer of a policy function as the input.
Recently, VAE-IW \cite{dittadi2021planning} obtains a compact binary representation by training a Binary-Concrete VAE offline.

Search-based approaches are in general not directly comparable against RL-based approaches
because the latter are given a significantly larger computational budgets than the former.
One exception is $\pi$-IW, a hybrid approach that requires $2\times 10^7$ interactions\footnote{Updated in the Arxiv version from $4\times 10^7$ in the ICAPS paper.}
to learn policies in Atari games.
This suggests a better middle ground between one extreme (RL) that requires billions of interactions,
and another (search) which does not improve in performance over time but is more sample efficient.
Given that $\pi$-IW requires a large training budget,
we hypothesized that learning a policy is significantly sample-inefficient.
By building off of VAE-IW, which performs offline representation learning that is fixed once trained,
and does not improve over time,
we aim to build an agent which \emph{does} learn over time, \emph{without} expensive policy learning.

To this end, we propose Olive (Online-VAE-IW),
a hybrid agent that collects new data and improves its state representation online
by combining Iterative Width with \emph{Active Learning} \cite{burr2012active}.
Active Learning is a group of approaches that optimize a data-dependent objective
by exploring, selecting, and adding new observations to the dataset.
Olive uses active learning in two senses:
First, it improves the search with a \emph{Multi-Armed Bandit} mechanism
that balances data collection (exploration) and reward-seeking behavior (exploitation)
based on expected rewards and their variances.
Second, it improves the state representation with \emph{Uncertainty Sampling} \cite{burr2012active} mechanism
that selects and adds novel screens between each planning episode based on the inaccuracy of the state representation.
Active learning collects reward information and screen information in the respective cases,
but there is a non-trivial interaction between them.
For example, reward-seeking behavior is also necessary for screen selection
because it is not cost effective to accurately learn the representations of states that are unpromising in terms of rewards.

Our online learning agent plays the game until a specified limit on ALE
simulator calls is reached.  Each time the game finishes (an \emph{episode}), the agent selects and adds a
subset of the observed game screens to a dataset, then retrains a Binary Concrete VAE neural
network that provides feature encodings of screens. Between each action taken in an episode, it constructs a
search tree by performing rollouts, and selects the most promising action (giving the
highest achievable reward in the expanded tree) at the root node using the information collected during the rollouts.
States encountered during a rollout are pruned
by Rollout-IW according to its novelty criteria. During a rollout, actions are selected by balancing
the exploration and the exploitation of rewards using the Best Arm Identification algorithm
called
Top Two Thompson sampling \cite{russo2020simple}.

We show that after only a small number of training episodes, our approach results in markedly
improved gameplay scores compared to state of the art width-based planning approaches and policy-based approaches with the same
or a smaller number of simulator calls.

\section{Background}
\label{sec:background}

We formalize the image-based ALE as a finite horizon MDP $P=\brackets{X,A,T,I,R,t^*,\gamma}$,
where
$X$ is a set of state observations (game screens) $\vx$,
$A$ is a set of actions $a$,
$T$ is a deterministic transition function,
$I\in X$ is an initial state,
$R: X, A, X\to \R$ is a reward function,
$t^*$ is a planning horizon,
and $\gamma$ is a discount factor.
A policy $\pi(\vx)=\Pr(a\mid \vx)$ is a categorical distribution of actions given a state,
and a score function $V_\pi(\vx)$ is the long-term reward of following $\pi$ from $\vx$, discounted by $\gamma$.
Our task is to find an optimal policy $\pi^*$ which maximizes $V_\pi^*(I)$.
Formally,
\begin{align}
 \vx_0 &= I,\ \vx_{t+1}= T(\vx_t,a_t),\\
 \pi^*&=\argmax_\pi V_\pi(I) = \argmax_\pi \sum_{t=0}^{t^*} \gamma^t R(\vx_t,a_t,\vx_{t+1}).\notag
\end{align}
In particular, we focus on a deterministic policy, i.e., a \emph{plan}, or an action sequence $\pi^*=(a_0,\cdots a_{t^*})$,
which is equivalent to $\Pr(a=a_t\mid \vx=\vx_t)=1, \Pr(a\not=a_t\mid \vx=\vx_t)=0$.
Additionally, we define an \emph{action-value} function
\begin{align}
 Q_{\pi}(\vx,a)&=R(\vx,a,\vx')+\gamma V_{\pi}(\vx'), \ \vx' = T(\vx,a)
\end{align}
which satisfies $V_{\pi^*}(\vx)=\max_a Q_{\pi^*}(\vx,a)$.
We will omit $\pi,\pi^*$ for brevity hereafter.

Following existing work, the agent is assumed to have access to the ALE simulator, and we assign a fixed simulation budget to the planning process and
force the agent to select an action each time the budget runs out.
We assume that observed states can be cached, and the
simulator can be set to any cached state during planning.
We discuss more details on the experimental setting in \refsec{sec:experiments}.

\subsection{Classical Novelty for Width-Based Planning}
\label{sec:iw}

Width-based planning is a group of methods that exploit the \emph{width} of a search problem,
originally developed for classical planning problems,
which are deterministic shortest path finding problems that are quite similar to MDPs in several aspects.
To simplify the discussion,
assume that each state $\vx\in X$ is represented by a subset of boolean atoms $F$, so that the total state space is $X = 2^{F}$.
A \emph{(conjunctive) condition} over the states can be represented as a subset $\vc\subseteq F$
where a state $\vx$ \emph{satisfies} $\vc$ (denoted as $\vx\satisfies \vc$) when $\vc\subseteq \vx$.
Then a unit-cost classical planning problem can be seen as
an MDP $\brackets{X,A,T,I,R,\infty,1}$ where $T$ is deterministic and,
given a goal condition $G$, $R(\vx,a,\vx')=0$ when $\vx'\satisfies G$, and $R(\vx,a,\vx')=-1$ otherwise.
Unlike typical STRIPS/PDDL-based classical planning \cite{FikesHN72}, $T$ could be a black-box in our setting.

Width in width-based planning \cite{lipovetzky2012width} is defined as follows.
Given a condition $\vc$, we call the optimal plan to reach a state $\vx\satisfies \vc$ to be the optimal plan to achieve $\vc$.
The \emph{width} $w(\vc)$ of $\vc$ is then the minimum size $|\vc'|$ of a condition $\vc'$
such that every optimal plan $\pi^*=(a_0,a_1\ldots,a_N)$ to achieve $\vc'$ is also an optimal plan to achieve $\vc$, and
that any prefix subsequence $(a_0,\ldots,a_n)$ of $\pi^*$ ($n\leq N$) is also an optimal plan for some condition of size $|\vc'|$.
The width of a classical planning problem is then $w(G)$, the width of the goal condition.
While the definition can be easily extended to categorical variables instead of boolean variables,
as seen in SAS+ formalism \cite{backstrom1995complexity},
we focus on boolean variables as they are easily interchangeable.

\citet{lipovetzky2012width} found that, in a majority of planning problems in International Planning Competition \cite{lopez2015deterministic} benchmarks,
each single goal atom $g\in G$ seen as a condition $\braces{g}$ tends to have a small width, typically below 2.
Based on this finding, they proposed \emph{Iterative Width} (IW),
a highly effective blind search algorithm that is driven by the intuition that
the \emph{original} problem
also has a small width.
IW performs a series of iterations over the width $w$ in increasing order.
The $w$-th iteration, denoted as $\iw(w)$, performs a breadth-first search that
enumerates optimal plans for every condition whose width is less than or equal to $w$.
$\iw(w)$ is guaranteed to solve a problem whose width is below $w$.
While it runs in time exponential in $w$, low-$w$ iterations run quickly and efficiently
by keeping track of a \emph{novelty} metric of each search state and pruning the states with novelty larger than $w$.
A state $\vx$ is called \emph{novel} with regard to $w$
if there exists a condition $\vc$ of size $w$ such that $\vx \satisfies \vc$,
and no other state $\vx'$ that has been seen so far satisfies $\vc$.
The novelty of a state is then defined as the minimum $w$ for which it is novel.

Novelty can be used like a heuristic function \cite{lipovetzky2017bwfs}
in traditional search algorithms such as Greedy Best First Search \cite{hoffmann01}.
However, unlike traditional heuristics it does not utilize the transition model or the goal condition, and its value is affected by the set of nodes already expanded by the search algorithm.
Let $\db$ be a CLOSE list; a database of states expanded during the search and their auxiliary information.
We denote the novelty of state $\vx$ by $n(\vx,\db)$ to emphasize the fact that it depends on the CLOSE list.
Initially, $\db=\emptyset$, but new states are added to $\db$ as more states are expanded.
To determine whether a new state $\vx$ is novel, check whether there is a condition $\vc$ satisfied by $\vx$ and not satisfied by any state in $\db$.
To implement a fast lookup, $\db$ is implemented with a set of conditions that are already achieved.
$\iw(w)$ terminates when $\db$ contains all conditions of size $w$ expressible under $F$.

\subsection{Rollout-IW (RIW)}
\label{sec:rollout-iw}

Width-based planning has been successfully applied to MDPs outside of classical planning. A boolean
encoding $\vz \in \B^F$ is computed for each state, where each element $\evz_j$ of the
encoding indicates that the $j$-th atom in $F$ is present in the state if and only if $\evz_j =
1$. A state is pruned if its novelty is larger than a specified width, and the
policy with highest reward in the pruned state space is determined.

To apply width-based planning to Atari games with states derived from screen pixels,
\citet{bandres2018planning} proposed to compute $\vz$ from a vector of pixels $\vx\in X$ using the
B-PROST feature set \cite{liang2016state}. However, even with width 1, the state space is too
large to plan over using breadth first search under the time budgets consistent with real-time planning.
To overcome this issue, they proposed Rollout-IW, which improves the anytime behavior of IW by
replacing the breadth-first search with depth-first rollouts. They define a CLOSE list $\db^d$ specific to each tree depth $d$
and use a depth-specific novelty $n(\vx,\db^{\leq D})$ for state $\vx$ at depth $D$. The depth-specific novelty is the
smallest $w$ such that for some condition $\vc$ with $|\vc| = w$, $\vx \satisfies \vc$ and $\vc$ is not
satisfied by any state in $\db^{d}$ with $d \leq D$. Rollouts are performed from the tree root,
selecting actions uniformly at random until a non-novel state is reached. Novelty is reevaluated for states
each time they are reached, since CLOSE lists at lower depths may change between different rollouts,
and a state is pruned if it is ever found to not be novel.

\subsection{VAE-IW}
\label{sec:vae-iw}
\label{sec:vae}

VAE-IW \cite{dittadi2021planning} extends RIW to learn encodings from screen images.
In a training stage, the game is run using RIW until a fixed number of screens are encountered and saved.
A Binary-Concrete Variational Autoencoder \cite{jang2017categorical,maddisonmt17} is trained on the saved data
to produce a binary encoding vector $\vz \in \B^F$ from screen pixels,
following work on learning a PDDL representation of image-based inputs \cite{Asai2018}.
After the training, the game is played using RIW, with features obtained by the encoder.

A VAE consists of an encoder network that returns $q_\phi(\vz \mid \vx)$,
a decoder network that returns $p_\theta(\vx \mid \vz)$,
and a prior distribution $p(\vz)$. The VAE is trained to maximize the \emph{evidence lower bound}
(ELBO) of the saved screens, computed as
\begin{align}
&\log p_\theta(\vx) \geq \text{ELBO}(\vx)\label{eq:elbo}\\
&= \E_{q_\phi(\vz | \vx)} \brackets{\log p_\theta(\vx \mid \vz)} - \KL\parens{ q_\phi(\vz \mid \vx) \Mid p(\vz) }.\notag
\end{align}

To obtain binary latent vectors, the latent variables are assumed to follow component-wise
independent Bernoulli distributions.
\begin{align}
  p(\vz) &= \prod_{j=1}^{F} \bern(0.5), & q_\phi(\vz \mid \vx) &= \prod_{j=1}^{F} \bern(\evmu_j)
\end{align}
where $\vmu$ are Bernoulli parameters that are obtained as the sigmoid of the output of the
encoder network. To obtain deterministic features for planning, $q_\phi(\vz \mid \vx)$ is
thresholded, using $\evz_j = 1$ if $\evmu_j > 0.9$ and 0 otherwise.

\section{Online Representation Learning for Atari}

Learning the features from data permits the encoding to be tailored to a specific game, but generating
a faithful encoding of a game is non-trivial, and using a static dataset is typically
insufficient. For example, \citet{dittadi2021planning} save screens that are reached by a
Rollout-IW agent using a hand-coded B-PROST feature set, and use those screens to train a
Binary-Concrete VAE to produce a game-specific encoding. However, in order for the
encoding to be representative of a game, the dataset must include screens from all visually distinct
parts, such as separate levels. When the B-PROST/Rollout-IW agent lacks a degree of
competence to reach level 2 and beyond, screens from the later levels are never included in the
dataset, thus VAE-IW cannot perform well in later levels.

To build an efficient online planning agent that simultaneously learns the representation,
we must tackle three challenges:
\textbf{(1)} \emph{How to automatically collect a diverse set of screens in the dataset},
which requires a metric that quantifies the diversity of screens.
\textbf{(2)} \emph{How to focus on improving the representations of states that return high rewards}.
This is because learning the representation of the states with low rewards may not be worthwhile
because the resulting agents will ultimately avoid such states.
Finally,
\textbf{(3)} \emph{How to keep the dataset size small.}
If the dataset gets too large, the computational effort for retraining between episodes becomes prohibitive.

To tackle these challenges,
we propose \emph{Olive} (Online-VAE-IW) (\refalgo{alg:Olive}),
an online planning and learning agent
that performs dynamic dataset refinement and retraining of the VAE between each search episode (\reflines{algl:aiw1}{algl:aiw2}).
It addresses the challenges above from three aspects:
\textbf{(1)} Pruning based on novelty,
\textbf{(2)} search guided by Bayesian estimates of future rewards, and
\textbf{(3)} active learning based on uncertainty sampling in games that change visually as the game progresses.
Olive consists of 3 nested loops:
\begin{enumerate}
 \item \emph{Episode} (\refline{algl:episode1}):
       A period which is started and ended by a game reset.
       Each reset could be triggered by a limit on the maximum number of actions in an episode,
       a limit on simulator calls (training phase only),
       or an in-game mechanism (e.g., beating the game, running out of lives).
       During training, each episode is followed by augmenting the dataset with $k$ screens
       observed in the episode and retraining the VAE. When evaluating Olive's performance
       after training, no additional screens are added.
 \item \emph{Planning and Acting} (\refline{algl:act1}):
       A period in which the agent makes a decision by searching the state space using a simulator.
       It ends by hitting a limit on the maximum number of simulator calls for the period.
       When it reaches the limit, the agent performs an action irreversibly.
       The agent chooses the action that achieved the largest reward $Q_i(\vx, a)$ on any rollout $i$ from
       the root node  (\refline{algl:act2}), and then truncates the search tree at the new root.
 \item \emph{Rollout} (\refline{algl:rollout1}):
       A lookahead that collects information using a black-box simulator.
       States are pruned if they are not novel (\refline{algl:termination}).
       Actions are chosen by \emph{Best Arm Identification} (BAI) algorithms (\refline{algl:bai}),
       based on statistics from previous rollouts.
       Each recursion into a deeper rollout obtains a new empirical $Q$-value (\reflines{algl:update1}{algl:update2})
       and updates the statistics (\reflines{algl:update3}{algl:update4}).
\end{enumerate}

\begin{algorithm}[tb]
  \caption{Training and evaluation of Olive}
  \label{alg:Olive}
\begin{algorithmic}[1]
  \Procedure{Olive-Experiment}{$X,A,T,I,R$}
    \State Initialize VAE, dataset $\X\from\emptyset$
    \For{$ep = 1, \dots, N_{ep}$} \Comment{Training loop}
      \State $\X \gets \X\cup \{ k$ screens observed in $\Call{Episode}{} \}$ \label{algl:aiw1} 
      \State Retrain VAE with $\X$ \label{algl:aiw2}
    \EndFor
    \For{$ep = 1, \dots, 10$} \Comment{Evaluation loop}
       \State \Call{Episode}{}
    \EndFor
  \EndProcedure

  \Procedure{Episode}{}\label{algl:episode1}
    \State $\vx_{root} \gets$ \Call{ResetGame}{}
    \While{$\vx_{root}$ is not terminal}\Comment{e.g., until dead}
      \State $\vx_{root} \gets$ \Call{PlanAndAct}{$\vx_{root}$}
    \EndWhile
    \State \Return observed screens
  \EndProcedure\label{algl:episode2}

  \Procedure{PlanAndAct}{$\vx_{root}$}\label{algl:act1}
    \State Initialize CLOSE list $\braces{\db^0\ldots}$
    \While{budget remaining} \Comment{Planning}
      \State \Call{rollout}{$\vx_{root},[],0$}
    \EndWhile
    \State \Return $T(\vx_{root}, \argmax_a \max_i Q_i(\vx_{root}, a))$
  \EndProcedure\label{algl:act2}

  \Procedure{rollout}{$\vx,\pi,d$}\label{algl:rollout1} \Comment{state, path, depth}
  \State $\vz \from \sigmoid(\encode(\vx)) > 0.9$
  \State \textbf{if} $\text{Parameters}[\pi]$ is not present: \Comment{new node}
  \State \qquad Update $\db^d$ using $\vx$, initialize $\text{Parameters}[\pi]$
  \State \textbf{if} $\vx$ is terminal $|$ budget exceeded $|$ $n(\vx,\db^{\leq d}) > w$:
  \State \qquad \Return 0 \label{algl:termination}
  \State $a \gets \text{Best Arm Identification}(\text{Parameters})$ \label{algl:bai}
  \State $\vx' \gets T(\vx,a),\ \pi' \gets \pi+[a]$
  \State $(n, \bar{\mu}, \bar{\sigma}^2) \gets \text{Parameters}[\pi']$
  \State $Q_{n+1}(\vx,a) \gets$
 \label{algl:update1}\\
         \qquad\qquad $R(\vx,a,\vx')+\gamma \function{rollout}(\vx',\pi',d+1)$
 \label{algl:update2}%
  \State $\bar{\mu}'\gets\frac{n\bar{\mu}+Q_{n+1}(\vx,a)}{n+1}$
 \label{algl:update3}
  \State $\bar{\sigma}^2{}'\gets \frac{(n+1) \bar{\sigma}^2}{n+2} + \frac{(Q_{n+1}(\vx,a)-\bar{\mu})(Q_{n+1}(\vx,a)-\bar{\mu}')}{n+2}$
 \label{algl:update4}
  \State $\text{Parameters}[\pi'] \gets (n+1, \bar{\mu}', \bar{\sigma}^2{}')$
  \State \Return $Q_{n+1}(\vx,a)$
  \EndProcedure\label{algl:rollout2}
\end{algorithmic}
\end{algorithm}

\subsection{Prior Distributions for Reward}

Let $Q(\vx,a)$ be a random variable for the cumulative reward of a rollout from $\vx$ starting with an action $a$,
and $Q_i(\vx,a)$ be a cumulative reward obtained from the $i$-th rollout launched from $\vx$ with $a$.
Assume that we have launched $n$ rollouts from $\vx$ with $a$ so far, and
we have a collection of data $Q_1(\vx,a) \ldots Q_n(\vx,a)$.
We define a hierarchical Bayesian model where
$p(Q(\vx,a))$ is a Gaussian distribution $\N(\mu,\sigma^2)$ with an unknown mean $\mu$ and a variance $\sigma^2$,
which are further modeled as follows \cite{gelman1995bayesian}:%
\footnote{
This particular modeling is equivalent to \emph{Normal-Gamma} distribution in other literature.
}
\begin{align}
 Q(\vx,a) \mid \mu, \sigma^2 \sim& \ \N(\mu, \sigma^2)           \label{eq:draw1}\\
 \mu \mid \sigma^2    \sim& \ \N(\bar{\mu}, \sigma^2 / n) \label{eq:draw2}\\
 \sigma^2             \sim& \ \text{Scaled-Inv}\chi^2(n+1, \bar{\sigma}^2)
\label{eq:normal-gamma}
\end{align}
where
$\bar{\mu}$ is the empirical mean $\bar{\mu}=\frac{1}{n}\sum_{i=1}^n Q_i(\vx,a)$ of the data,
and $\bar{\sigma}^2$ is a \emph{regularized} variance
$\bar{\sigma}^2=\frac{1}{n+1} (\sigma_0 + \sum_i (Q_i(\vx,a)-\bar{\mu})^2)$ where we use $\sigma_0=0.2$ as a prior (hyperparameter),
$\text{Scaled-Inv}\chi^2$ is a Scaled Inverse $\chi^2$ distribution.
Initializing $\bar{\sigma}^2$ to a non-zero value $\sigma_0$ when no rollouts are made ($n=0$) and
using $n+1$ as the divisor instead of $n$ is called a pseudo-count method,
which avoids \emph{zero-frequency problems}, i.e.,
when it is the first time to launch a rollout from the current node,
we avoid computing $\text{Scaled-Inv}\chi^2$ with $\bar{\sigma}^2 = 0$, which is undefined.

This Bayesian statistical modeling can properly account for the uncertainty of the observed $Q(\vx,a)$.
The key idea is that the true mean/variance of $Q(\vx,a)$
differ from the empirical mean/variance of $Q(\vx,a)$ observed during finite rollouts.
Since we do not know the true mean/variance, we use distributions to model which values of mean/variance are more likely.
This distribution becomes more precise as more observations are made, e.g.,
the variance $\sigma^2 / n$ of $\mu$ decreases as the divisor $n$ increases (more observations).

To incorporate this Bayesian model, it is only necessary to store and update
$n, \bar{\mu}, \bar{\sigma}^2$ incrementally on each search node (\reflines{algl:update3}{algl:update4})
without storing the sequence $Q_1(\vx,a)\ldots Q_n(\vx,a)$ explicitly.
See Appendix \refsec{sec:incremental-update} for the proof of these updates.
Note the $n+2$ being used as the divisor for the variance due to the pseudo-count, which adds 1 to the number of samples.
In all our experiments, we initialize the parameters to $n=0$, $\bar{\mu}=0$, $\bar{\sigma}^2 = 0.2$.
We refer to this set of parameters as $\parens{n,\bar{\mu},\bar{\sigma}^2}=\text{Parameters}[\pi']$,
where $\pi$ is an action sequence that reached $\vx$ and $\pi' = \pi + [a]$.

Although Atari environments are deterministic,
note that $Q(\vx,a)$ still have uncertainty due to the depth-first algorithm
which does not systematically search all successors.

\subsection{Best Arm Identification}

While Rollout-IW and VAE-IW have focused on uniform action selection within the pruned state space,
we recognize that more advanced action selection is fully compatible with width-based planning. We
use the distributions over $Q$ to apply a Best-Arm Identification (BAI) strategy at each node \cite{audibert2010best,karnin2013almost}.

BAI is a subclass of the multi-armed bandit (MAB) problem \cite{gittins2011multi};
a sequential decision making problem in which many unknown distributions (levers) are present, and one is
selected to be sampled (pulled) from at each decision. MABs have frequently been studied with a minimum cumulative
regret (CR) objective, which aims to maximize the expected reward of the samples \emph{obtained during the experimentation}.
For example, UCB1 \cite{auer2002finite} is the theoretical basis of the
widely used UCT algorithm for MCTS \cite{kocsis2006bandit}, and is derived under a CR objective.
In contrast to CR, BAI draws samples to minimize the \emph{simple regret}, i.e.,
to maximize the expected rewards of a single decision (acting) made \emph{after} the experimentation.
BAI is thus more appropriate for our problem setting, as we wish to maximize the performance of acting
and not the cumulative performance of rollouts.
A number of works have recognized BAI as a more natural objective for tree search and have
proposed algorithms based on BAI
\cite{cazenave2014sequential,kaufmann2017monte} or hybrids of BAI and
cumulative regret minimization \cite{pepels2014minimizing}. To our knowledge,
these search procedures have not been applied within a state space
pruned by width-based planning.
\citet{shleyfman2016blind} also point out that simple regret is better suited for this purpose.
However, their Racing p-IW algorithm is not explicitly based on BAI.

In this paper, we consider Top-Two Thompson Sampling \citep[TTTS]{russo2020simple} as a BAI algorithm, and additionally consider
UCB1, uniform random action selection, and maximum empirical mean for reference.
Note that Rollout-IW and VAE-IW use uniform sampling for rollouts.

\textbf{Top-Two Thompson Sampling (TTTS)}
is an algorithm that is based on Thompson Sampling (TS).
TS runs as follows:
For each $a$,
we first sample $\sigma^2$ (\refeq{eq:normal-gamma}).
Next, using $n$, $\bar{\mu}$, and the sampled value of $\sigma^2$, we sample $\mu$ (\refeq{eq:draw2}).
We similarly sample $Q(\vx,a)$ (\refeq{eq:draw1}).
TS returns $\hat{a}=\argmax_{a} Q(\vx,a)$ as the best action.
TTTS modifies the action returned by TS
with probability $\alpha$ (a hyperparameter, 0.5 in our experiments). TTTS discards action $\hat{a}$,
then continues running TS until it selects a different action $a\not=\hat{a}$, which is returned as a solution.
TTTS is shown to provide a better exploration than TS.  Since we use an improper prior at $n=0$, we first sample each action once at a new node.

\textbf{Upper Confidence Bound (UCB1)}
selects the action which maximizes the metric
$\bar{\mu}_a + \sqrt{2\log N / n_a}$
where $\bar{\mu}_a$ is the statistic $\bar{\mu}$ for action $a$, $n_a$ is the count $n$ for $a$, and $N=\sum_{a} n_a$ is the sum of counts across actions.

\subsection{Updating the Screen Dataset}
\label{sec:bi}

Finally, we update the dataset $\X$ by adding a fixed number $k$ of new observations after each episode.
In principle, all screens that are observed during
an episode could be added to a dataset to further train the VAE.
However, this approach quickly exhausts the physical memory and prolongs the training time between episodes.

We address this issue through two strategies. The first approach is to select the $k$ screens
at random from those observed during the episode. This approach is more likely to include screens from
later in the game as the agent improves, incorporating visual changes in the game into the learned representation.
We refer to this approach as \emph{Passive Olive}.

However, most of the new data are duplicates, or are quite close to data points already in the dataset.
For example, 
the initial state and states nearby may be added to the dataset multiple times,
and are likely to have similar appearances.
As a result, the dataset contains more screens for the states near the initial states
than those for the states deep down the search tree, resulting in an \emph{imbalanced dataset}.
Not only do these duplicate screens lack new information for the VAE to learn,
but the imbalance also potentially prevents VAEs from learning new features,
because machine learning models assume i.i.d. samples
and tend to ignore rare instances in an imbalanced dataset \cite{wallace2011class}.

To tackle this challenge we employ \emph{Uncertainty Sampling} \cite{burr2012active},
a simple active learning strategy which selects data that the current model is most uncertain about.
Retraining the model with those data points is expected to improve the accuracy of the model for those newly added inputs.
In uncertainty sampling, we select $k$ screens
$\X_{\text{chosen}}$ from among the newly collected screens $\X_{\text{new}}$ with the lowest total
probability under the current model:
\begin{equation*}
  \argmin_{\X_{\text{chosen}} \subset \X_{\text{new}}, |\X_{\text{chosen}}| = k} \ \sum_{\vx \in \X_{\text{chosen}}} \log p_{\theta}(\vx).
\end{equation*}

In Olive, $p_\theta(\vx)$ represents the VAE.
Since $p_\theta(\vx)$
is unknown, we approximate it with its lower bound, i.e., the ELBO that is also the loss function of the VAE (\refeq{eq:elbo}).
Therefore, given a new set of observations $\X_{\text{new}}$,
we compute their loss values using the VAE obtained in the previous iteration,
then select the screens with the top-$k$ highest loss values (minimum ELBO) as $\X_{\text{chosen}}$.
Images with high loss values are images that the VAE is uncertain about, and which will be improved
when added to the training set.
We call the resulting configuration of Olive as \emph{Active Olive}.

\section{Empirical Evaluations}
\label{sec:experiments}

We evaluated Olive and existing approaches on a compute cluster running an
AMD EPYC 7742 processor with nVidia Tesla A100 GPUs.
We used 55 Atari 2600 games with
a problem-dependent number of actions \cite[the \emph{minimal action} configuration]{bellemare2013arcade}
and the risk-averse reward setting \cite{bandres2018planning}.
For every configuration,
we trained the system with 5 different random seeds,
then evaluated the result for 10 episodes (playthroughs) for 50 evaluations total.
Following the existing work \cite{junyent2019deep,junyent2021hierarchical,dittadi2021planning},
we use the discount factor of $\gamma=0.99$ (rather than 0.995 in \cite{bandres2018planning}) in \refline{algl:update2}.
However, note that the reported final scores are undiscounted sums of rewards.

\textbf{Resource Constraints:}
We repeated the setup of previous work \cite{lipovetzky2012width,bandres2018planning,dittadi2021planning}
where each action is held for 15 frames, which we refer to as taking a single action. 
We define a single \emph{simulator call} as an update to the simulator when taking a single action.
Improving the learned state representation requires screens to be observed by agents with increasing
performance that can, for example, reach further in the game and experience later levels. In order to
allow Olive to play a larger number of episodes during training, we limit the number
of actions in each episode to 200. During evaluation, an episode ends if the game is not already finished
once 18000 actions have been performed from the initial state (because some games can be played indefinitely).

To evaluate the sample efficiency, we limit the total number of simulator calls allowed for the training, or the \emph{total training budget}.
Approaches that contain machine learning with IW include
VAE-IW, $\pi$-IW, and HIW \cite{dittadi2021planning,junyent2019deep,junyent2021hierarchical},
but they use widely different budgets.
The latter two use $4\times 10^7$ or $2\times 10^7$ simulator calls,
while VAE-IW does not specify such a limit due to how they collect the data.
VAE-IW collects 15000 images in total
by indefinitely playing the game with B-PROST features, and randomly sampling 5 images in the search tree between taking each action. To normalize by budget, we modify VAE-IW to stop when it exhausts the budget,
and sample $k=15000$ images out of all past observed images using reservoir sampling.
The total budget is $10^5$ simulator calls,
which is adequate because we use only 15000 images in total.
The VAE is trained for 100 epochs, following the README file of the public source code of VAE-IW (not specified in the paper).

Olive, in contrast, plays the first episode with B-PROST features
and collects $k=500$ images from the observed screens.
After the first iteration, it extends the dataset by $k=500$ images in each episode,
but by performing Rollout-IW using the features learned at the end of the previous episode.
If the training budget is exhausted before completing 30 episodes (15000 images),
the remaining images are sampled from the entire set of screens seen in the final episode.

Finally, to achieve ``almost real-time'' performance proposed in \cite{bandres2018planning},
we limit the computational resources that can be spent between taking each action, or the ``budget''.
An issue with the ``almost real-time'' setting of some existing work is that
the budget between each action is limited by runtime.
This lacks reproducibility because the score is affected by the performance of the compute hardware.
Following \cite{junyent2019deep,junyent2021hierarchical},
we instead limit the budget by the number of simulator calls, which is set to 100.

\textbf{Evaluation Criteria:}
Since scores in Atari domains have varying magnitudes, we cannot compare them in different environments directly.
Instead, we count the number of ``wins/losses'' between configurations.
The wins are the number of environments where one configuration outperformed another with a statistical significance.
To test the significance, we used Mann-Whitney's $U$ test with $p < 0.05$.

\subsection{Ablation Study}
\label{sec:ablation}

We perform an ablation study of Olive to understand the effect of each improvement added to VAE-IW.
We test the effects of VAE training parameters, action selection strategy in rollouts, and active learning of the dataset.
\reftbl{tab:summary-vae} shows a summary of these comparisons.

\paragraph{VAE-IW + Temperature Annealing:}
Our VAEs are identical to those described in \cite{dittadi2021planning},
consisting of 7 convolutional layers in the encoder and the decoder.
However, they made a questionable choice for training Binary-Concrete VAE:
They did not anneal the temperature $\tau$ \cite{jang2017categorical} during the training,
keeping the value low ($\tau=0.5$).
A BC-VAE becomes slower to train when $\tau$ is small
because the activation function becomes closer to a step function, with small gradients away from the origin.
We evaluated an exponential schedule $\tau=\tau_{\max}e^{-Ct}$ for an epoch $t$
and an appropriate constant $C$
which anneals $\tau=\tau_{\max}=5.0$ to $\tau=0.5$ at the end of the training ($t=100$).

We compared the total number of significant wins of $\tau_{\max} = 5.0$ against $\tau_{\max} = 0.5$,
thereby enabling/disabling annealing.
$\tau_{\max} = 5.0$ won against $\tau_{\max} = 0.5$ in 17 domains, while it lost in 14 domains.
From this result, we conclude that annealing causes a significant improvement in score.
We call this variation VAE-IW+ann.

\paragraph{VAE-IW + ann + BAI Rollout:}
We next evaluated the effect of adding active learning in action selection to VAE-IW+ann using BAI.
\reftbl{tab:VAEIWBanditExperiment} shows the wins between BAI algorithms,
including uniform (baseline), UCB1, TTTS,
and max (a greedy baseline that selects the action with the maximum empirical mean reward).
The results indicate that TTTS significantly outperforms uniform and max
by winning in 10 and 9 games, and losing in 5 and 2 games, respectively.
While UCB1 has 6 wins and 2 losses against TTTS,
it did not outperform uniform decisively (4 wins and 3 losses).
Since TTTS outperforms uniform more reliably,
we conclude that VAE-IW+ann+TTTS is the best approach without online dataset updates.

\begin{table}[tb]
\centering
\begin{tabular}{r|cccc}
 \diagbox[width=5em,height=1.5em]{win}{loss} & uniform & max & UCB1 & TTTS \\
\midrule
uniform       & -                & 9      & 3      & 5 \\
max           & 5 (-4)           & -      & 4      & 2 \\
UCB1          & 4 (+1)           & 8 (+4) & -      & 6 \\
\textbf{TTTS} & \textbf{10} (+5) & 9 (+7) & 2 (-4) & - \\
\bottomrule
\end{tabular}
\caption{
Win/loss comparisons for different BAI algorithms in VAE-IW+ann.
$X (\pm Y)$ indicates $X$ wins and $Y$ wins minus the losses.
TTTS outperformed uniform and max in significantly more domains.
UCB1 won against TTTS in more domains in direct comparison,
but failed to win decisively against uniform (4 wins, 3 losses).
}
\label{tab:VAEIWBanditExperiment}
\end{table}

\paragraph{Offline Learning vs. Passive Olive vs. Active Olive:}
Finally, we test the effect of online representation learning by
comparing Passive and Active Olive to offline approaches in \reftbl{tab:summary-vae}.
Active Olive is more effective than Passive Olive (6-to-4 against Passive),
and wins more against offline learning
(Active wins 7-to-5 while Passive wins 7-to-6, against the best offline approach, VAE-IW+ann+TTTS.)

The impact of Active Olive is most
important for games whose visual features change significantly at higher scores.
For example,
in BankHeist, JamesBond, Pitfall, 
and WizardOfWor, the
background and enemy designs change once an agent reaches far enough
in the game, while in Amidar 
 new enemy designs and color
swaps occur.
Montezuma's Revenge has multiple rooms,
and the maximum score of Olive reached 2500 without policy learning,
which outperforms 540 reported by 2BFS \citep{lipovetzky2015classical}.
However,
the impact of active online learning is game dependent.
In games whose visual features are rather static, the VAE trained by VAE-IW can be sufficient.

We also see Passive Olive could outperform Active Olive in some games.
This is due to the weakness of uncertainty sampling that,
while it seeks for new screens,
it does not consider the similarity between the $k$ selected screens,
i.e., selected screens can be visually similar to each other and can skew the dataset distribution.
The more sophisticated Active Learning methods (e.g., mutual information maximization)
address this issue, but we leave the extension for future work.
Instead,
to assess the best possible scores achievable with online learning,
we analyzed a hypothetical, oracular portfolio, labeled as PortfolioOlive,
which counts the domains in which either ActiveOlive or PassiveOlive wins against the baselines,
and domains in which both lost against the baselines.
As expected, this approach even more significantly outperformed existing baselines.

\begin{table*}[tbp]
  \centering
\begin{adjustbox}{width=0.8\linewidth}
\begin{tabular}{>{\raggedleft}p{6em}|cccccc|c}
 \diagbox[width=6em,height=1.5em]{win}{loss} & RIW & VAE-IW & +ann & +ann+TTTS & \textbf{PassiveOlive} & \textbf{ActiveOlive} & \textbf{PortfolioOlive} \\
\midrule
RIW                     & -                 & 8                 & 10              & 7              & 4      & 6 & 4  \\
VAE-IW                  & 31 (+23)          & -                 & 14              & 13             & 11     &13 & 10 \\
+ann                    & 31 (+21)          & 17 (+3)           & -               & 5              & 5      & 7 & 4  \\
+ann+TTTS               & 38 (+31)          & 20 (+7)           & 10 (+5)         & -              & 6      & 5 & 3  \\
\textbf{PassiveOlive}            & 36 (+32)          & 21(+10)           & 6(+1)           & 7 (+1)         & -      & 4 & -  \\
\textbf{ActiveOlive}             & 35 (+29)          & 20(+7)            & 7($\pm$0)       & 7 (+2)         & 6 (+2) & - & -  \\
\midrule
\textbf{PortfolioOlive} & {38 (+34)} & {26 (+16)} & {10(+6)} & {9(+6)} & -      & - & -  \\
\bottomrule
\end{tabular}
\end{adjustbox}
\caption{
Ablation study on Olive comparing wins/losses.
In each $X (\pm Y)$, $X$ indicates the number of wins, and $Y$ indicates the number of wins minus losses.
Adding annealing and a Bayesian BAI (TTTS) active learning significantly improves the performance.
Active representation learning
improved the performance over passive representation learning,
but they are complementary.
Their hypothetical, oracular portfolio (the best online configuration) shows future prospects.
}
\label{tab:summary-vae}
\end{table*}

\begin{table*}[tbp]
\centering
\begin{adjustbox}{width=\linewidth}
\begin{tabular}{r|rr|rrr|rrrr|r}
\toprule
 & human & VAEIW & RIW & VAEIW & Olive & Olive & $\pi$-IW & DQN & EfficientZero & Olive  \\[-0.21em]
train/plan budget &  & NA/0.5s & 0/100 & $10^5$/100 & $10^5$/100 & $10^5$/100 & $2\times 10^7$/100 & $5\times 10^7$/NA & $10^5$/(187.5) & sec./act. \\[-0.21em]
\midrule
Alien & 6875 & 7744 & 6539 & \textbf{7535} & 5450 & \textbf{5450} & 3969.8 & 3069 & 1140.3 & 0.41 \\[-0.21em]
Amidar & 1676 & 1380.3 & 537 & 928 & \textbf{1390} & \textbf{1390} & 950.4 & 739.5 & 101.9 & 0.4 \\[-0.21em]
Assault & 1496 & 1291.9 & 1053 & 1374 & \textbf{1477} & 1477 & 1574.9 & \textbf{3359} & 1407.3 & 0.3 \\[-0.21em]
Asterix & 8503 & 999500 & 919580 & \textbf{999500} & \textbf{999500} & \textbf{999500} & 346409.1 & 6012 & 16843.8 & 0.31 \\[-0.21em]
Asteroids & 13157 & 12647 & 2820 & 33799 & \textbf{41213} & \textbf{41213} & 1368.5 & 1629 & - & 0.33 \\[-0.21em]
Atlantis & 29028 & 1977520 & 62538 & \textbf{2327026} & 2266404 & \textbf{2266404} & 106212.6 & 85641 & - & 0.39 \\[-0.21em]
Bank Heist & 734.4 & 289 & 241 & 219 & \textbf{301} & 301 & \textbf{567.2} & 429.7 & 361.9 & 0.34 \\[-0.21em]
Battle zone & 37800 & 115400 & 42640 & 33760 & \textbf{67660} & 67660 & \textbf{69659.4} & 26300 & 17938 & 0.38 \\[-0.21em]
Beam rider & 5775 & 3792 & \textbf{4435} & 2786 & 3785 & 3785 & 3313.1 & \textbf{6846} & - & 0.38 \\[-0.21em]
Berzerk & - & 863 & 720 & \textbf{847} & 702 & 702 & \textbf{1548.2} & - & - & 0.33 \\[-0.21em]
Bowling & 154.8 & 54.4 & 26 & \textbf{64} & \textbf{64} & \textbf{64} & 26.3 & 42.4 & - & 0.33 \\[-0.21em]
Boxing & 4.3 & 89.9 & \textbf{100} & 99 & 98 & 98 & \textbf{99.9} & 71.8 & 44.1 & 0.44 \\[-0.21em]
Breakout & 31.8 & 45.7 & 6 & \textbf{58} & 53 & 53 & 92.1 & 401.2 & \textbf{406.5} & 0.39 \\[-0.21em]
Centipede & 11963 & 428451.5 & 103983 & \textbf{623726} & 129967 & \textbf{129967} & 126488.4 & 8309 & - & 0.36 \\[-0.21em]
Chopper command & 9882 & 4190 & 13490 & 14842 & \textbf{37248} & \textbf{37248} & 11187.4 & 6687 & 1794 & 0.4 \\[-0.21em]
Crazy climber & 35411 & 901930 & 95056 & 308970 & \textbf{735826} & \textbf{735826} & 161192 & 114103 & 80125.3 & 0.32 \\[-0.21em]
Demon attack & 3401 & 285867.5 & 23882 & 128101 & \textbf{128817} & \textbf{128817} & 26881.1 & 9711 & 13298 & 0.37 \\[-0.21em]
Double dunk & -15.5 & 8.6 & 5 & 4 & \textbf{8} & \textbf{8} & 4.7 & -18.1 & - & 0.39 \\[-0.21em]
ElevatorAction & - & 40000 & 84688 & \textbf{113890} & 83826 & \textbf{83826} & - & - & - & 0.36 \\[-0.21em]
Enduro & 309.6 & 55.5 & 1 & \textbf{65} & 25 & 25 & \textbf{506.6} & 301.8 & - & 0.41 \\[-0.21em]
Fishing derby & 5.5 & -20 & -36 & \textbf{-30} & -33 & -33 & \textbf{8.9} & -0.8 & - & 0.46 \\[-0.21em]
Freeway & 29.6 & 5.3 & 7 & \textbf{8} & 7 & 7 & 0.3 & \textbf{30.3} & 21.8 & 0.49 \\[-0.21em]
Frostbite & 4335 & 259 & 416 & 270 & \textbf{509} & \textbf{509} & 270 & 328.3 & 313.8 & 0.39 \\[-0.21em]
Gopher & 2321 & 8484 & 4740 & 6925 & \textbf{10073} & 10073 & \textbf{18025.9} & 8520 & 3518.5 & 0.34 \\[-0.21em]
Gravitar & 2672 & 1940 & 535 & 1358 & \textbf{1655} & 1655 & \textbf{1876.8} & 306.7 & - & 0.32 \\[-0.21em]
Ice hockey & 0.9 & 37.2 & 30 & 34 & \textbf{37} & \textbf{37} & -11.7 & -1.6 & - & 0.39 \\[-0.21em]
James bond 007 & 406.7 & 3035 & 338 & 2206 & \textbf{7637} & \textbf{7637} & 43.2 & 576.7 & 459.4 & 0.36 \\[-0.21em]
Kangaroo & 3035 & 1360 & 796 & 884 & \textbf{1192} & 1192 & 1847.5 & \textbf{6740} & 962 & 0.39 \\[-0.21em]
Krull & 2395 & 3433.9 & 3428 & \textbf{4086} & 3818 & 3818 & \textbf{8343.3} & 3805 & 6047 & 0.41 \\[-0.21em]
Kung-fu master & 22736 & 4550 & \textbf{6488} & 5764 & 4874 & 4874 & \textbf{41609} & 23270 & 31112.5 & 0.39 \\[-0.21em]
Montezuma & 4367 & 0 & 0 & 0 & \textbf{58} & \textbf{58} & 0 & 0 & - & 0.38 \\[-0.21em]
Ms. Pac-man & 15693 & 17929.8 & 15951 & \textbf{19614} & 16890 & \textbf{16890} & 14726.3 & 2311 & 1387 & 0.37 \\[-0.21em]
Name this game & 4076 & 17374 & 14109 & 14711 & \textbf{15804} & \textbf{15804} & 12734.8 & 7257 & - & 0.36 \\[-0.21em]
Phoenix & - & 5919 & 5770 & \textbf{6592} & 6165 & \textbf{6165} & 5905.1 & - & - & 0.32 \\[-0.21em]
Pitfall! & - & -5.6 & -40 & -50 & \textbf{-12} & \textbf{-12} & -214.8 & - & - & 0.37 \\[-0.21em]
Pong & 9.3 & 4.2 & -6 & \textbf{-5} & \textbf{-5} & -5 & -20.4 & 18.9 & \textbf{20.6} & 0.36 \\[-0.21em]
Private eye & 69571 & 80 & 57 & \textbf{1004} & 400 & 400 & 452.4 & \textbf{1788} & 100 & 0.4 \\[-0.21em]
Q*bert & 13455 & 3392.5 & 1543 & \textbf{7165} & 6759 & 6759 & \textbf{32529.6} & 10596 & 15458.1 & 0.41 \\[-0.21em]
Riverraid & 13513 & 6701 & \textbf{7063} & 6538 & 6303 & 6303 & - & \textbf{8316} & - & 0.39 \\[-0.21em]
Road Runner & 7845 & 2980 & 13268 & 21720 & \textbf{36636} & 36636 & \textbf{38764.8} & 18257 & 18512.5 & 0.4 \\[-0.21em]
Robotank & 11.9 & 25.6 & \textbf{64} & 35 & 61 & \textbf{61} & 15.7 & 51.6 & - & 0.45 \\[-0.21em]
Seaquest & 20182 & 842 & 913 & 1785 & \textbf{2469} & 2469 & \textbf{5916.1} & 5286 & 1020.5 & 0.36 \\[-0.21em]
Skiing & 1652 & -10046.9 & -29950 & -29474 & \textbf{-28143} & -28143 & \textbf{-19188.3} & - & - & 0.38 \\[-0.21em]
Solaris & - & 7838 & \textbf{7808} & 5155 & 7378 & \textbf{7378} & 3048.8 & - & - & 0.45 \\[-0.21em]
Space invaders & 1652 & 2574 & 2761 & 2682 & \textbf{3301} & \textbf{3301} & 2694.1 & 1976 & - & 0.34 \\[-0.21em]
Stargunner & 10250 & 1030 & 2232 & 3164 & \textbf{5234} & 5234 & 1381.2 & \textbf{57997} & - & 0.3 \\[-0.21em]
Tennis & -8.9 & 4.1 & -14 & \textbf{8} & 6 & \textbf{6} & -23.7 & -2.5 & - & 0.4 \\[-0.21em]
Time pilot & 5925 & 32840 & 13824 & \textbf{27946} & 25180 & \textbf{25180} & 16099.9 & 5947 & - & 0.34 \\[-0.21em]
Tutankham & 167.7 & 177 & \textbf{158} & 147 & 149 & 149 & \textbf{216.7} & 186.7 & - & 0.35 \\[-0.21em]
Up’n down & 9082 & 762453 & \textbf{834351} & 729061 & 741694 & \textbf{741694} & 107757.5 & 8456 & 16095.7 & 0.54 \\[-0.21em]
Venture & 1188 & 0 & 0 & \textbf{6} & 0 & 0 & 0 & \textbf{380} & - & 0.35 \\[-0.21em]
Video pinball & 17298 & 373914.3 & 319596 & 437727 & \textbf{464096} & 464096 & \textbf{514012.5} & 42684 & - & 0.45 \\[-0.21em]
Wizard of wor & 4757 & 199900 & 142582 & 175066 & \textbf{197662} & \textbf{197662} & 76533.2 & 3393 & - & 0.37 \\[-0.21em]
Yars’ revenge & - & 96053.3 & 69344 & 87940 & \textbf{90778} & 90778 & \textbf{102183.7} & - & - & 0.34 \\[-0.21em]
Zaxxon & 9173 & 15560 & 6496 & 10472 & \textbf{10594} & 10594 & \textbf{22905.7} & 4977 & - & 0.36 \\[-0.21em]
\bottomrule
\end{tabular}
\end{adjustbox}
\caption{
Comparison of the average scores. 
Best scores in each group in \textbf{bold}.
We also included human and VAE-IW scores from \cite{dittadi2021planning} as a reference.
Scores of existing work are
from the cited papers except $\pi$-IW (based on its Arxiv manuscript, as per authors' request).
Hyphens indicate missing data.
EfficientZero's planning budget is adjusted for frameskip 15.
The rightmost column shows the average runtime of Olive between actions in seconds.
}
\label{tab:main}
\end{table*}

\subsection{Main Results: Olive vs. $\pi$-IW vs. EfficientZero}

Finally, \reftbl{tab:main} provides individual average scores of ActiveOlive across 55 Atari domains,
alongside scores achieved by Rollout-IW and VAE-IW.
In terms of average scores, ActiveOlive outperforms VAE-IW in \textbf{32-to-20},
and Rollout-IW in \textbf{42-to-11}.
Olive is also \textbf{best in class in 30 domains}.
Standard errors and maximum scores are in \reftbls{tab:max}{tab:complete} in the appendix.
As a reference, we added human scores
and VAE-IW scores from \cite{dittadi2021planning} which uses hardware-specific 0.5 second planning budget and an unknown amount of training budget.

Next, to understand the impact of a good representation relative to the impact of policy-learning,
\reftbl{tab:main} also compares ActiveOlive against $\pi$-IW \cite{junyent2019deep}, DQN \cite{dqn} and EfficientZero \cite{ye2021mastering}.
All results are obtained from the cited papers,
except $\pi$-IW which is based on its Arxiv manuscript, as per its authors' request.
$\pi$-IW is a width-based approach guided by a neural policy function which
discretizes its intermediate layer as the feature vector for width-based search.
This is an interesting comparison because $\pi$-IW has a significantly larger total training budget ($2\times 10^7$)
compared to ActiveOlive ($10^5$), while operating under the same planning budget of 100.
DQN is a state of the art model-free RL approach.
EfficientZero is a recent state of the art in model-based data-efficient RL
trained on a subset of Atari domains under the same $10^5$ environment interactions.
It uses a frameskip of 4 and 50 MCTS simulations per acting, instead of 15 and 100 in Olive,
thus is allowed to see the screen $15/4=3.75$ times more frequently,
and is given about 1.8x more planning budget per second (equivalent to $50\cdot \frac{15}{4} = 187.5$ simulations per acting in frameskip 15).
\textbf{ActiveOlive outperforms $\pi$-IW by 30-to-22, DQN by 31-to-17, EfficientZero by 18-to-7.}
VAE-IW also outperforms EfficientZero by 15-to-10.
These wins/losses are counted by comparing the average results without the $U$-test
due to the lack of the data on each run.
Olive has average runtime per action of 0.38 sec.\@,
which is closely above the real-time performance (15 frames, 0.25 sec.)
that would be achievable by optimization (e.g., C++) and faster hardware.

\section{Related Work}

Incorporating Multi-Armed Bandit strategies to address
exploration-exploitation trade off in search problems
has a long history.
Indeed, the initial evaluation of the planning-based setting of Atari
was done based on UCB1 \cite{bellemare2013arcade}.
However, from a theoretical standpoint,
UCB1 is not appropriate in \emph{simple regret} minimization scenario like online planning and acting.
To our knowledge,
we are the first to use a Bayesian BAI algorithm (TTTS) in MCTS.
A limited body of work has used frequentist BAI in
MCTS (e.g. Successive Halving in \cite{cazenave2014sequential}),
but not within width-pruned search or the Atari domain.

$\pi$-IW \cite{junyent2019deep} learns a $Q$-function
which is then combined with Rollout-IW by replacing the uniform action selection
with a softmax of $Q$ values.
However, it does not explicitly consider the uncertainty of the current estimate.
In other words, the $Q$-function in $\pi$-IW provides a point estimate (mean)
of $Q(\vx,a)$, rather than its distributional estimate.
Thus, while we focused on knowledge-free uninformed search as the basis
of our method, our contribution of uncertainty-aware Bayesian methods is
orthogonal to policy learning approaches.
Extending $\pi$-IW to Bayesian estimates is future work.

Olive is limited to learning the state representation,
requiring simulator interactions for rollouts.
To further improve the sample efficiency,
future work includes learning the environment dynamics,
similar to model-based RL \cite{muzero}
but without expensive, sample-inefficient policy learning.
Learning a PDDL/STRIPS model \cite{Asai2020}
and leveraging off-the-shelf heuristic functions that provide a search guidance without learning
is an interesting avenue of future work.

\section{Conclusions}

In this paper, we proposed Olive, an online extension of VAE-IW, which
obtains a compact state representation of the screen images of Atari games
with a variational autoencoder.
Olive incrementally improves the quality of the learned representation
by actively searching for states that are both new and rewarding,
based on well-founded Bayesian statistical principles.
Experiments showed that our agent is competitive against
a state of the art width-based planning approach that was trained with more than 100 times larger training budget,
demonstrating Olive's high sample efficiency.

\fontsize{9pt}{10pt}\selectfont

\clearpage

\appendix
\section{Appendix}
\subsection{Binary-Concrete Variational Autoencoders}

A variational autoencoder (VAE) is an unsupervised representation learning method
that transforms an observed vector $\vx$ into a latent vector $\vz$,
typically of a smaller dimension than $\vx$.
Let $p(\vx)$ be the unknown ground-truth distribution of a data point $\vx$ and
$q(\vx)$ be a \emph{dataset distribution} of the finite dataset $\X$ at hand.
$q(\vx)\not=p(\vx)$ because 
\begin{align}
 q(\vx) =
 \left\{
 \begin{array}{cc}
  1/|\X| & \vx \in \X\\
  0 & \vx \not\in \X
 \end{array}
 \right.
\end{align}
while $p(\vx)$ contains all possible points outside the dataset.
Additionally, let $p_\theta(\vx)$ be a distribution represented by a sufficiently expressive neural network with weights $\theta$
which, for some $\theta=\theta^*$, $p_{\theta^*}(\vx)=p(\vx)$.
The training of a VAE is mathematically justified by the principle of \emph{maximum likelihood estimation} (MLE)
which, given a dataset $\X$ (thus $q(\vx)$),
repeatedly samples a value $\vx$ from $q(\vx)$,
then optimizes weights $\theta$ by maximizing the probability $p_\theta(\vx)$ of observing such a data point.
It is assumed that the MLE parameters converge to the ground truth, i.e., $\theta^*=\argmax_\theta p_\theta(\vx)=\argmax_\theta \log p_\theta(\vx)$.
(Logarithm preserves the maxima.)

Since maximizing $\log p_\theta(\vx)$ is \#P-hard \cite{dagum1993approximating,roth1996hardness,dagum1997optimal},
VAEs approximate it by maximizing a lower bound called the ELBO \cite{kingma2013auto}.
ELBO is formulated by defining a \emph{variational distribution} $q_\phi(\vz \mid \vx)$
parameterized by neural network with weights $\phi$, as we describe below.

As stated in the main.text, a VAE consists of an encoder network that returns $q_\phi(\vz \mid \vx)$ given $\vx$,
a decoder network that returns $p_\theta(\vx \mid \vz)$ given $\vz$,
and a prior distribution $p(\vz)$. The VAE is trained to maximize the \emph{evidence lower bound}
(ELBO) defined as
\begin{align}
&\log p_\theta(\vx) \geq \text{ELBO}(\vx)\\
&= \E_{q_\phi(\vz | \vx)} \brackets{\log p_\theta(\vx \mid \vz)} - \KL\parens{ q_\phi(\vz \mid \vx) \Mid p(\vz) },\notag
\end{align}
where $\KL\parens{q_\phi(\vz \mid \vx) \Mid p(\vz)}$ denotes the Kullback-Leibler divergence (KL divergence) from $p(\vz)$ to $q_\phi(\vz \mid \vx)$.

$q_\phi(\vz \mid \vx)$ is both a posterior distribution
(i.e., a distribution of latent variables $\vz$ given observed variables $\vx$,
 or a distribution \emph{after/post} the experimentation),
and a variational distributions
(i.e., an approximation of the true posterior distribution $p(\vz \mid \vx)$).
The prior distribution $p(\vz)$
(i.e., a distribution assumed by \emph{default}, or a distribution assumed \emph{before/prior to} the experimentation)
is fixed and does not depend on neural network parameters.
When training, ELBO is estimated by a single-sample monte-carlo, i.e.,
all expectations $\E$ are replaced by drawing a random sample only once,
i.e., running the network once for $p_\theta(\vx \mid \vz)$ and $q_\phi(\vz \mid \vx)$.

To obtain binary latent vectors, the latent variables are assumed to follow component-wise
independent Bernoulli distributions,
\begin{align}
  p(\vz) &= \prod_{j=1}^{F} \bern(m), & q_\phi(\vz \mid \vx) &= \prod_{j=1}^{F} \bern(\evmu_j)
\end{align}
where $m, \vmu$ are Bernoulli parameters.
The assumption of component-wise independent latent vectors is known as the \emph{mean-field
  assumption}. Typically, for Bernoulli distributions, the prior is selected to be $m=0.5$ \cite{jang2017categorical}.
$\vmu$ are obtained as $\vmu=\sigmoid(\vl)$,
where $\vl$ is an unactivated output $\vl=\encode(\vx)$ of the encoder, which is called a \emph{logit}.
For Bernoulli $q_\phi$ and $p$, KL divergence is computed as
\begin{align}
 \KL\parens{ q_\phi \Mid p }
 =\textstyle \sum_j \parens{\evmu_j \log \frac{\evmu_j}{m} + (1-\evmu_j)\log \frac{1-\evmu_j}{1-m}}.
\end{align}

To reconstruct $\vx$ using $p_\theta(\vx \mid \vz)$, the decoder acts on samples drawn from $q_\phi(\vz \mid \vx)$.
Since the use of Bernoulli distributions makes the network non-differentiable,
samples are taken from its continuous relaxation during the training, called a Binary-Concrete distribution
\cite{jang2017categorical,maddisonmt17}, denoted as $\BC$.
Sampling from $\BC$ takes the logit $\vl$ as the input and is obtained
from a \emph{temperature} $\tau$ and a sample from a logistic distribution:
\begin{align}
 \textstyle
 \BC(\vl)=\sigmoid\parens{\frac{\vl+\function{Logistic}(0,1)}{\tau}}.
\end{align}
Sampling from $\BC$ converges to a Bernoulli distribution in the limit $\tau\rightarrow 0$.
i.e., $\BC(\vl)\to\function{step}(\vl)=(\vl<0)\,?\,0:1$.

\paragraph{$\beta$-VAE}

A modification to VAE training that improves the quality of generated $\vx$ in certain domains is
the $\beta$-VAE framework \cite{higgins2017beta}, where the following modified objective is maximized
\begin{equation} \E_{q_\phi(\vz | \vx)} \brackets{\log p_\theta(\vx \mid \vz)} - \beta \ \KL\parens{
q_\phi(\vz \mid \vx) \Mid p(\vz) }.
\end{equation} Use of $\beta$ controls the relative importance of reconstruction of training data
and keeping the distribution of $\vz$ close to the prior. A value of $\beta = 0.0001$ is used in
VAE-IW \cite{dittadi2021planning}. Since our VAEs replicate those used by
\citeauthor{dittadi2021planning}, we have used the same value.

\subsection{Incremental Update of Parameters}
\label{sec:incremental-update}

Given a sequence of data $x_1\ldots x_n$ and a new data $x_{n+1}$,
their empirical mean $\bar{\mu}$ and variance $\bar{\sigma^2}$ can be updated to $\bar{\mu}'$ and $\bar{\sigma}^2{}'$ as follows.

Proof for the means:

\begin{align}
 \bar{\mu} &= \frac{\sum_{i=1}^{n} x_i}{n}. \\
 \bar{\mu}' &= \frac{\sum_{i=1}^{n+1} x_i}{n+1} = \frac{\parens{\sum_{i=1}^{n} x_i} + x_{n+1}}{n+1}= \frac{n\bar{\mu}+x_{n+1}}{n+1}.
\end{align}

Proof for the variance:

\begin{align}
 \bar{\sigma}^2 &= \frac{\sum_{i=1}^{n} \parens{x_i-\bar{\mu}}^2}{n}. \\
 \bar{\sigma}^2{}' &= \frac{\sum_{i=1}^{n+1} \parens{x_i-\bar{\mu}'}^2}{n+1}. \\
 (n+1)\bar{\sigma}^2{}'-n\bar{\sigma}^2 &= \sum_{i=1}^{n+1} \parens{x_i-\bar{\mu}'}^2-\sum_{i=1}^{n} \parens{x_i-\bar{\mu}}^2\\
 =\parens{x_{n+1}-\bar{\mu}'}^2+\sum_{i=1}^{n}& \parens{\parens{x_i-\bar{\mu}'}^2-\parens{x_i-\bar{\mu}}^2}\\
 =\parens{x_{n+1}-\bar{\mu}'}^2+\sum_{i=1}^{n}
 &\parens{\parens{x_i-\bar{\mu}'}+\parens{x_i-\bar{\mu}}}\\
 &\parens{\parens{x_i-\bar{\mu}'}-\parens{x_i-\bar{\mu}}}\\
 =\parens{x_{n+1}-\bar{\mu}'}^2+\sum_{i=1}^{n}& \parens{2x_i-(\bar{\mu}'+\bar{\mu})}\parens{\bar{\mu}-\bar{\mu}'}\\
 =\parens{x_{n+1}-\bar{\mu}'}^2+& \parens{2n\bar{\mu}-n\bar{\mu}'-n\bar{\mu}}\parens{\bar{\mu}-\bar{\mu}'}\\
 =\parens{x_{n+1}-\bar{\mu}'}^2+& \parens{n\bar{\mu}-n\bar{\mu}'}\parens{\bar{\mu}-\bar{\mu}'}\\
 =\parens{x_{n+1}-\bar{\mu}'}^2+& \parens{n\bar{\mu}'-n\bar{\mu}}\parens{\bar{\mu}'-\bar{\mu}}.
\end{align}

We use the fact that $n\bar{\mu}'-n\bar{\mu}=x_{n+1}-\bar{\mu}'$. Then

\begin{align}
 =&\parens{x_{n+1}-\bar{\mu}'}^2+ \parens{x_{n+1}-\bar{\mu}'}\parens{\bar{\mu}'-\bar{\mu}}\\
 =&\parens{x_{n+1}-\bar{\mu}'} \parens{x_{n+1}-\bar{\mu}'+\bar{\mu}'-\bar{\mu}}\\
 =&\parens{x_{n+1}-\bar{\mu}'} \parens{x_{n+1}-\bar{\mu}}.
\end{align}

Therefore

\begin{align}
 (n+1)\bar{\sigma}^2{}'-n\bar{\sigma}^2 &= \parens{x_{n+1}-\bar{\mu}'} \parens{x_{n+1}-\bar{\mu}}.\\
 \bar{\sigma}^2{}' &= \frac{n\bar{\sigma}^2 + \parens{x_{n+1}-\bar{\mu}'} \parens{x_{n+1}-\bar{\mu}}}{n+1}.
\end{align}

\subsection{VAE-IW experiments}

In order to highlight the difference in training approach between VAE-IW and
Olive variants, \refalgo{alg:VAEIW} provides pseudocode for offline learning
with VAE-IW.

\begin{algorithm}[htb]
  \caption{Pseudocode for VAE-IW experiments}
  \label{alg:VAEIW}
\begin{algorithmic}[1]
  \Procedure{VAE-IW-Experiment}{$X,A,T,I,R$}
    \State Initialize VAE, dataset $\X\from\emptyset$
    \While{within training budget} \Comment{Training loop}
      \State $\X \gets \X\ \cup $ \Call{Episode}{}
    \EndWhile
    \State $\X \gets \function{shuffle}(\X)[:k]$ \Comment{Random Sampling}
    \State Train VAE with $\X$
    \For{$ep = 1, \dots, 10$} \Comment{Evaluation loop}
       \State \Call{Episode}{}
    \EndFor
  \EndProcedure
\end{algorithmic}
\end{algorithm}

\subsection{The list of iteration limits in Olive}

Olive has three hyperparameters that limit the number of iterations in each algorithm.
We summarize our hyperparameter choices in addition to budgets imposed during experiments
in \reftbl{tab:limits}.
Each of these parameters are also described in detail by the help
messages of the source code that we submitted in the supplemental material.

\begin{table}[htb]
\centering
\begin{tabular}{l|r}
\toprule
 Total training budget & $10^5$ \\
 Training epochs per VAE training & 100 \\
 Max. training episodes & 30 \\
 Max. acting per episode & 200 \\
 Max. simulator interactions between acting & 100 or 200 \\
\bottomrule
\end{tabular}
\caption{Budgets and hyperparameters used during experiments on Olive and baselines.
}
\label{tab:limits}
\end{table}

\section{Additional Tables}
\label{sec:additional}

\reftbl{tab:max} shows the maximum scores achieved by Rollout-IW, VAE-IW, and Olive.
In \reftbl{tab:max},
Olive managed to achieve the maximum score of 2500 in Montezuma's revenge,
and is an only agent that achieved a positive score in this game in our experiments, though this may have occurred purely by chance.

\reftbl{tab:complete} shows the mean gameplay scores alongside standard errors derived from the 50 evaluations of each game,
which could not be added to the main paper due to space limitations.

\reftbl{tab:evaluation-settings} shows a survey of the various evaluation settings in existing work,
which vary between papers, for future reference.

\begin{table}[htbp]
\centering
\begin{adjustbox}{height=30em}
\begin{tabular}{r|r|rrr}
\toprule
 & human  & RIW & VAEIW  & Olive  \\
\midrule
Alien & 6875 & 15700 & \textbf{16260} & 11960 \\
Amidar & 1676 & 1205 & 2915 & \textbf{3447} \\
Assault & 1496 & 1831 & \textbf{2697} & 2674 \\
Asterix & 8503 & \textbf{999500} & \textbf{999500} & \textbf{999500} \\
Asteroids & 13157 & 6590 & 94820 & \textbf{135680} \\
Atlantis & 29028 & 124400 & \textbf{2376500} & 2310100 \\
Bank Heist & 734.4 & \textbf{1325} & 570 & 1305 \\
Battle zone & 37800 & 121000 & 111000 & \textbf{709000} \\
Beam rider & 5775 & \textbf{13620} & 9990 & 12660 \\
Berzerk & - & \textbf{1740} & 1440 & 1720 \\
Bowling & 154.8 & 44 & 93 & \textbf{94} \\
Boxing & 4.3 & \textbf{100} & \textbf{100} & \textbf{100} \\
Breakout & 31.8 & 20 & 296 & \textbf{350} \\
Centipede & 11963 & 227780 & \textbf{1309954} & 213233 \\
Chopper command & 9882 & 73200 & 73900 & \textbf{229500} \\
Crazy climber & 35411 & 152600 & 633900 & \textbf{1238900} \\
Demon attack & 3401 & 85270 & \textbf{133790} & 132055 \\
Double dunk & -15.5 & 16 & 12 & \textbf{18} \\
ElevatorAction & - & \textbf{225100} & 208800 & 198600 \\
Enduro & 309.6 & 7 & \textbf{210} & 89 \\
Fishing derby & 5.5 & \textbf{3} & -13 & -3 \\
Freeway & 29.6 & 12 & \textbf{20} & 9 \\
Frostbite & 4335 & 4190 & 320 & \textbf{4450} \\
Gopher & 2321 & 11820 & 11420 & \textbf{19900} \\
Gravitar & 2672 & 2000 & 4500 & \textbf{4800} \\
Ice hockey & 0.9 & 41 & 48 & \textbf{49} \\
James bond 007 & 406.7 & 650 & 19000 & \textbf{23150} \\
Kangaroo & 3035 & 1400 & 1800 & \textbf{2800} \\
Krull & 2395 & \textbf{4840} & 4780 & 4530 \\
Kung-fu master & 22736 & 13400 & \textbf{13800} & 9200 \\
Montezuma & 4367 & 0 & 0 & \textbf{2500} \\
Ms. Pac-man & 15693 & \textbf{32681} & 29891 & 26101 \\
Name this game & 4076 & \textbf{26310} & 23010 & 23010 \\
Phoenix & 9.3 & 7220 & \textbf{12900} & 12720 \\
Pitfall! & - & \textbf{0} & \textbf{0} & \textbf{0} \\
Pong & - & 6 & \textbf{9} & 8 \\
Private eye & 69571 & 269 & 15100 & \textbf{15200} \\
Q*bert & 13455 & 6000 & 18175 & \textbf{18550} \\
Riverraid & 13513 & 9190 & 8820 & \textbf{9260} \\
Road Runner & 7845 & 48600 & 48800 & \textbf{156900} \\
Robotank & 11.9 & \textbf{82} & 68 & 81 \\
Seaquest & 20182 & 2460 & \textbf{6170} & 5780 \\
Skiing & 1652 & -27553 & -22778 & \textbf{-21304} \\
Solaris & - & 18400 & \textbf{18540} & 16400 \\
Space invaders & - & 4190 & 4565 & \textbf{4780} \\
Stargunner & 10250 & 6400 & 12900 & \textbf{26400} \\
Tennis & -8.9 & -7 & \textbf{19} & 18 \\
Time pilot & 5925 & 24100 & \textbf{92600} & 42900 \\
Tutankham & 167.7 & \textbf{245} & 185 & 185 \\
Up’n down & 9082 & 858400 & 971190 & \textbf{999990} \\
Venture & 1188 & 0 & \textbf{300} & 0 \\
Video pinball & 17298 & 999520 & \textbf{999999} & 999955 \\
Wizard of wor & 4757 & \textbf{199900} & \textbf{199900} & \textbf{199900} \\
Yars’ revenge & - & 89523 & \textbf{133916} & 118738 \\
Zaxxon & 9173 & 22300 & \textbf{25800} & \textbf{25800} \\
\midrule
Best in group & & 14 & 22 & \textbf{28} \\
\bottomrule
\end{tabular}
\end{adjustbox}
\caption{
Comparison of the maximum scores of Rollout-IW, VAE-IW, and Olive,
with human scores for a reference.
Best scores in \textbf{bold}.
}
\label{tab:max}
\end{table}

\begin{table*}
  \centering
\begin{adjustbox}{height=30em}
\begin{tabular}{r|rr|rrr}
\toprule
       &       & VAEIW      & RIW  & VAEIW & Olive \\
budget & human & 0.5s       & 100  & 100   & 100   \\
\midrule
Alien & 6875 & 7744 & 6539 $\pm$ 460 & 7535 $\pm$ 481 & 5450 $\pm$ 333 \\
Amidar & 1676 & 1380.3 & 537 $\pm$ 30 & 928 $\pm$ 60 & \textbf{1390 $\pm$ 86} \\
Assault & 1496 & 1291.9 & 1053 $\pm$ 43 & 1374 $\pm$ 57 & \textbf{1477 $\pm$ 56} \\
Asterix & 8503 & 999500 & 919580 $\pm$ 38717 & 999500 $\pm$ 0 & \textbf{999500 $\pm$ 0} \\
Asteroids & 13157 & 12647 & 2820 $\pm$ 162 & 33799 $\pm$ 2714 & \textbf{41213 $\pm$ 3945} \\
Atlantis & 29028 & 1977520 & 62538 $\pm$ 2310 & 2327026 $\pm$ 3220 & 2266404 $\pm$ 3965 \\
Bank Heist & 734.4 & 289 & 241 $\pm$ 29 & 219 $\pm$ 19 & \textbf{301 $\pm$ 31} \\
Battle zone & 37800 & 115400 & 42640 $\pm$ 3135 & 33760 $\pm$ 2583 & \textbf{67660 $\pm$ 15571} \\
Beam rider & 5775 & 3792 & 4435 $\pm$ 433 & \textbf{2786 $\pm$ 253} & 3785 $\pm$ 400 \\
Berzerk & - & 863 & 720 $\pm$ 39 & 847 $\pm$ 40 & 702 $\pm$ 39 \\
Bowling & 154.8 & 54.4 & 26 $\pm$ 1 & 64 $\pm$ 1 & \textbf{64 $\pm$ 1} \\
Boxing & 4.3 & 89.9 & 100 $\pm$ 0 & \textbf{99 $\pm$ 0} & 98 $\pm$ 0 \\
Breakout & 31.8 & 45.7 & 6 $\pm$ 1 & 58 $\pm$ 6 & 53 $\pm$ 7 \\
Centipede & 11963 & 428451.5 & 103983 $\pm$ 7036 & 623726 $\pm$ 60845 & 129967 $\pm$ 6333 \\
Chopper command & 9882 & 4190 & 13490 $\pm$ 1709 & 14842 $\pm$ 1835 & \textbf{37248 $\pm$ 6093} \\
Crazy climber & 35411 & 901930 & 95056 $\pm$ 5016 & 308970 $\pm$ 18756 & \textbf{735826 $\pm$ 37635} \\
Demon attack & 3401 & 285867.5 & 23882 $\pm$ 2428 & 128101 $\pm$ 2389 & \textbf{128817 $\pm$ 837} \\
Double dunk & -15.5 & 8.6 & 5 $\pm$ 1 & 4 $\pm$ 1 & \textbf{8 $\pm$ 1} \\
ElevatorAction & - & 40000 & 84688 $\pm$ 15449 & 113890 $\pm$ 14081 & 83826 $\pm$ 13518 \\
Enduro & 309.6 & 55.5 & 1 $\pm$ 0 & 65 $\pm$ 6 & 25 $\pm$ 3 \\
Fishing derby & 5.5 & -20 & -36 $\pm$ 2 & -30 $\pm$ 1 & -33 $\pm$ 2 \\
Freeway & 29.6 & 5.3 & 7 $\pm$ 0 & 8 $\pm$ 1 & 7 $\pm$ 0 \\
Frostbite & 4335 & 259 & 416 $\pm$ 102 & 270 $\pm$ 2 & \textbf{509 $\pm$ 122} \\
Gopher & 2321 & 8484 & 4740 $\pm$ 354 & 6925 $\pm$ 504 & \textbf{10073 $\pm$ 445} \\
Gravitar & 2672 & 1940 & 535 $\pm$ 50 & 1358 $\pm$ 131 & \textbf{1655 $\pm$ 133} \\
Ice hockey & 0.9 & 37.2 & 30 $\pm$ 1 & 34 $\pm$ 1 & \textbf{37 $\pm$ 1} \\
James bond 007 & 406.7 & 3035 & 338 $\pm$ 22 & 2206 $\pm$ 640 & \textbf{7637 $\pm$ 1136} \\
Kangaroo & 3035 & 1360 & 796 $\pm$ 46 & 884 $\pm$ 55 & \textbf{1192 $\pm$ 63} \\
Krull & 2395 & 3433.9 & 3428 $\pm$ 63 & 4086 $\pm$ 78 & 3818 $\pm$ 68 \\
Kung-fu master & 22736 & 4550 & 6488 $\pm$ 326 & \textbf{5764 $\pm$ 325} & 4874 $\pm$ 235 \\
Montezuma & 4367 & 0 & 0 $\pm$ 0 & 0 $\pm$ 0 & \textbf{58 $\pm$ 50} \\
Ms. Pac-man & 15693 & 17929.8 & 15951 $\pm$ 906 & 19614 $\pm$ 1013 & 16890 $\pm$ 800 \\
Name this game & 4076 & 17374 & 14109 $\pm$ 580 & 14711 $\pm$ 537 & \textbf{15804 $\pm$ 489} \\
Phoenix & 9.3 & 5919 & 5770 $\pm$ 42 & 6592 $\pm$ 364 & 6165 $\pm$ 272 \\
Pitfall! & - & -5.6 & -40 $\pm$ 12 & -50 $\pm$ 18 & \textbf{-12 $\pm$ 5} \\
Pong & - & 4.2 & -6 $\pm$ 1 & -5 $\pm$ 1 & \textbf{-5 $\pm$ 1} \\
Private eye & 69571 & 80 & 57 $\pm$ 28 & 1004 $\pm$ 505 & 400 $\pm$ 302 \\
Q*bert & 13455 & 3392.5 & 1543 $\pm$ 195 & 7165 $\pm$ 742 & 6759 $\pm$ 712 \\
Riverraid & 13513 & 6701 & 7063 $\pm$ 144 & \textbf{6538 $\pm$ 150} & 6303 $\pm$ 144 \\
Road Runner & 7845 & 2980 & 13268 $\pm$ 2032 & 21720 $\pm$ 2281 & \textbf{36636 $\pm$ 2810} \\
Robotank & 11.9 & 25.6 & 64 $\pm$ 1 & \textbf{35 $\pm$ 2} & 61 $\pm$ 1 \\
Seaquest & 20182 & 842 & 913 $\pm$ 70 & 1785 $\pm$ 167 & \textbf{2469 $\pm$ 160} \\
Skiing & 1652 & -10046.9 & -29950 $\pm$ 49 & -29474 $\pm$ 546 & \textbf{-28143 $\pm$ 367} \\
Solaris & - & 7838 & 7808 $\pm$ 727 & \textbf{5155 $\pm$ 539} & 7378 $\pm$ 643 \\
Space invaders & - & 2574 & 2761 $\pm$ 140 & 2682 $\pm$ 162 & \textbf{3301 $\pm$ 182} \\
Stargunner & 10250 & 1030 & 2232 $\pm$ 162 & 3164 $\pm$ 287 & \textbf{5234 $\pm$ 774} \\
Tennis & -8.9 & 4.1 & -14 $\pm$ 0 & 8 $\pm$ 1 & 6 $\pm$ 1 \\
Time pilot & 5925 & 32840 & 13824 $\pm$ 672 & 27946 $\pm$ 2152 & 25180 $\pm$ 1256 \\
Tutankham & 167.7 & 177 & 158 $\pm$ 4 & \textbf{147 $\pm$ 2} & 149 $\pm$ 3 \\
Up’n down & 9082 & 762453 & 834351 $\pm$ 1862 & \textbf{729061 $\pm$ 48689} & 741694 $\pm$ 46001 \\
Venture & 1188 & 0 & 0 $\pm$ 0 & 6 $\pm$ 6 & 0 $\pm$ 0 \\
Video pinball & 17298 & 373914.3 & 319596 $\pm$ 36518 & 437727 $\pm$ 55805 & \textbf{464096 $\pm$ 45553} \\
Wizard of wor & 4757 & 199900 & 142582 $\pm$ 7255 & 175066 $\pm$ 8463 & \textbf{197662 $\pm$ 1612} \\
Yars’ revenge & - & 96053.3 & 69344 $\pm$ 1475 & 87940 $\pm$ 1996 & \textbf{90778 $\pm$ 1774} \\
Zaxxon & 9173 & 15560 & 6496 $\pm$ 743 & 10472 $\pm$ 1207 & \textbf{10594 $\pm$ 1026} \\
\midrule
Best in group &  &  & 8 & 20 & \textbf{30} \\
\bottomrule
\end{tabular}
\end{adjustbox}
\caption{
(\reftbl{tab:main} with standard errors of the means.)
Comparison of the average scores of Rollout-IW, VAE-IW, $\pi$-IW and Olive,
with human scores and VAE-IW scores from \cite{dittadi2021planning} for a reference.
Missing data are indicated by hyphens.
Best scores in \textbf{bold}.
Each numbers are $\text{mean}\pm\text{stderr}$, where $\text{stderr}=\text{stdev}/\sqrt{50}$.
}
\label{tab:complete}
\end{table*}

\begin{table*}[htbp]
\centering
\begin{adjustbox}{height=\textheight}
\rotatebox{90}{
\begin{tabular}{|l|c|c|c|c|c|c|c|c|c|c|c|c|c|c|c|c|}
\hline
paper & \multicolumn{ 3}{c|}{\shortcite{bellemare2013arcade}} & \shortcite{bellemare2015arcade} & \multicolumn{ 2}{c|}{\shortcite{lipovetzky2015classical}} & \shortcite{shleyfman2016blind} & \shortcite{jinnai2017learning} & \multicolumn{ 3}{c|}{\shortcite{bandres2018planning}} & \shortcite{junyent2019deep} & \shortcite{junyent2021hierarchical} & \shortcite{dittadi2021planning} & \multicolumn{ 2}{c|}{ours} \\ \hline
name & RL & BrFS & UCT & Brute & UCT & IW, 2BFS & Racing pIW & DASA & IW & IW & RIW & $\pi$-IW & $\pi$-IW+, HIW & VAE-IW & VAE-IW & Olive \\ \hline
\textbf{General settings} & \multicolumn{ 16}{c|}{} \\ \hline
state &  & RAM & RAM & RAM & RAM & RAM & RAM & RAM & RAM & b-prost & b-prost & b-prost & b-prost & VAE & VAE & VAE \\ \hline
novelty &  &  &  &  &  & 1 & 1 & 1 & 1 & 1 & depth & dynamic* & ? & depth & depth & depth \\ \hline
discount factor & .999 & .999 & .999 & .999 & .995 & .995 & .995 & .995 & .995 & .995 & .995 & .99 & .99 & .99 & .99 & .99 \\ \hline
frameskip &  & 5 & 5 & 5 & 5 & 5 & 5 & 5 & 5 & 15 & 15 & 15 & 15 & 15 & 15 & 15 \\ \hline
actions / episode &  & 18k & 18k & 18k & 18k & 18k & 18k & 18k & 18k & 18k & 18k & 18k & 18k & 15k & 18k & 18k \\ \hline
rewards &  &  &  &  &  &  &  &  &  &  & RA,RAS & ? & ? & RA & RA & RA \\ \hline
\textbf{Training} &  &  &  &  &  &  &  &  &  &  &  &  &  &  &  &  \\ \hline
Total frames &  &  &  &  &  &  &  &  &  &  &  & 600M & 300M &  & 1.5M & 1.5M \\ \hline
Total training budget &  &  &  &  &  &  &  &  &  &  &  & 40M & 20M &  & 0.1M & 0.1M \\ \hline
Training dataset size &  &  &  &  &  &  &  &  &  &  &  &  &  & 15k & 15k & 15k \\ \hline
Total episodes & 5k &  &  &  &  &  &  &  &  &  &  &  &  &  &  &  \\ \hline
\textbf{Planning, Rollout} &  &  &  &  &  &  &  &  &  &  &  &  &  &  &  &  \\ \hline
runtime per action &  &  &  &  &  &  &  &  &  & \multicolumn{ 2}{c|}{0.25, 0.5, 32} &  &  & 0.5, 32 &  &  \\ \hline
max frames / action &  & 133k &  &  & 150k & 150k & 150k, 10k &  & 150k &  &  & 1500 & 1500 &  & 1500 & 1500 \\ \hline
max nodes / action &  &  &  &  & 150k,30k & 30k & 30k, 2k &  & 30k &  &  & 100 & 100 &  & 100 & 100 \\ \hline
max nodes / rollout &  &  & 300 &  & 300 & 1500 & 1500 &  & 1500 &  &  &  &  &  & 100 & 100 \\ \hline
max rollouts / action &  &  & 500 &  & 500 &  &  &  &  &  &  &  &  &  &  &  \\ \hline
frameskip in rollout &  &  & 1 &  & 1,5 &  &  &  &  &  &  &  &  &  &  &  \\ \hline
\textbf{Others} &  &  &  &  &  &  &  &  &  &  &  &  &  &  &  &  \\ \hline
Random seeds & 30 & 10 & 10 &  & 10 & 10 & 10 & 10 & 10 & 10 & 10 & 5 & 10 & 10 & 5 & 5 \\ \hline
Eval episodes & 500 &  &  &  &  &  &  &  &  &  &  & 10 & 10 & 10 & 10 & 10 \\ \hline
exploration strategy &  &  & UCB1 & e-greedy & UCB1 &  &  &  &  &  &  &  &  &  &  & TTTS \\ \hline
policy function &  &  & DP &  &  &  &  &  &  &  & uniform & learned & learned & uniform & uniform & DP \\ \hline
value function &  &  &  &  &  &  &  &  &  &  &  &  & learned &  &  &  \\ \hline
\end{tabular}
}
\end{adjustbox}
\caption{
Survey of the evaluation settings in existing work.
(*): $\pi$-IW discretizes the last hidden layer of the policy network to obtain features.
}
\label{tab:evaluation-settings}
 \end{table*}

\end{document}